\documentclass[10pt,twocolumn,letterpaper]{article}
\usepackage{cvpr}
\usepackage{times}
\usepackage{epsfig}
\usepackage{graphicx}
\usepackage{amsmath}
\usepackage{amssymb}
\usepackage{algorithm}
\usepackage{algorithmic}
\usepackage{subfigure}
\usepackage{stfloats}
\usepackage[T1]{fontenc}
\usepackage{authblk}
\usepackage[utf8]{inputenc}
\usepackage[pagebackref=true,breaklinks=true,letterpaper=true,colorlinks,bookmarks=false]{hyperref}

\cvprfinalcopy % *** Uncomment this line for the final submission

\def\cvprPaperID{8044} % *** Enter the CVPR Paper ID here

\ifcvprfinal\fi
\begin{document}

%%%%%%%%% TITLE
\title{Towards the Automation of Deep Image Prior}
%\author[1]{Qianwei Zhou\thanks{zhouqianweischolar@gmail.com}}
%\author[1]{Chen Zhou \thanks{hcqylym@gmail.com}}
%\author[1]{Haigen Hu \thanks{hghu@zjut.edu.cn}}
%\author[1]{Yuhang Chen \thanks{xiaobai456123@gmail.com}}
%\author[2]{Shengyong Chen \thanks{sy@ieee.org}}
%\author[1]{Xiaoxin Li\thanks{mordekai@zjut.edu.cn}}
%
%\renewcommand{\thefootnote}{\fnsymbol{footnote}}
%\affil[1]{Zhejiang University of Technology}
%\affil[2]{Tianjin University of Technology}

\author{
        Qianwei~Zhou,
        Chen~Zhou,
        Haigen~Hu,
        Yuhang~Chen,
        Shengyong~Chen,
        Xiaoxin~Li% <-this % stops a space
\thanks{This work was supported in part the National Natural Science Foundation of China under Grants 61802347, U1509207, and 61374094, the Natural Science Foundation of Zhejiang Province under Grants LY18F030025 and LY18F020031. Q. Zhou,C. Zhou, H. Hu, C. Chen, and X. Li (corresponding author) are with College of Computer Science and Technology, Zhejiang University of Technology, Hangzhou 310023, China
(e-mail: zhouqianweischolar@gmail.com, hcqylym@gmail.com, hghu@zjut.edu.cn, xiaobai456123@gmail.com, mordekai@zjut.edu.cn). S. Chen is with the School of Computer Science and Engineering, Tianjin University of Technology, Tianjin 300384, China, and with Zhejiang University of Technology, Hangzhou 310023, China (e-mail: sy@ieee.org).}%
        }

\maketitle
%\thispagestyle{empty}

%%%%%%%%% ABSTRACT

\begin{abstract}
Single image inverse problem is a notoriously challenging ill-posed problem that aims to restore the original image from one of its corrupted versions. Recently, this field has been immensely influenced by the emergence of deep-learning techniques. Deep Image Prior (DIP) offers a new approach that forces the recovered image to be synthesized from a given deep architecture. While DIP is quite an effective unsupervised approach, it is deprecated in real-world applications because of the requirement of human assistance.  
\par
In this work, we aim to find the best-recovered image without the assistance of humans by adding a stopping criterion, which will reach maximum when the iteration no longer improves the image quality. More specifically, we propose to add a pseudo noise to the corrupted image and measure the pseudo-noise component in the recovered image by the orthogonality between signal and noise. The accuracy of the orthogonal stopping criterion has been demonstrated for several tested problems such as denoising, super-resolution, and inpainting, in which 38 out of 40 experiments are higher than 95\%. 
\end{abstract}

%%%%%%%%% BODY TEXT
\section{Introduction}
Image inverse problem center around the recovery of an unknown image $x$ based on given corrupted measurement $y$. It is an ill-posed problem because a specific corrupted image $y$ can correspond to a crop of possible high-quality images. The problem has been extensively explored in the past several decades while deep convolutional neural networks (ConvNets) currently set the state-of-the-art \cite{ulyanov2018deep}, such as denoising \cite{cruz2018nonlocality}, or single-image super-resolution \cite{hui2018fast}. The commonly suggested and very effective path to the inverse problem is as follows: Given many example of pairs of an original image and its corrupted version, one could learn a deep network to match the degraded image to its source \cite{mataev2019deepred}, for example, \cite{yang2019deep, zhang2018ffdnet, lefkimmiatis2018universal, plotz2018neural, liu2018non, park2018srfeat, zhang2018image, wang2018fully, tao2018scale, gao2019dynamic}.

Ulyanov \etal~\cite{ulyanov2018deep} proposed a new strategy, namely Deep Image Prior (DIP), for a single image inverse problem where common strategies are on longer feasible because only one corrupted image (without the original image) is available for model training. Mataev \etal~\cite{mataev2019deepred} further improved the performance of the DIP by adding an extra regularization (Regularization by Denoising). 

Although Ulyanov \etal~\cite{ulyanov2018deep} and Mataev \etal~\cite{mataev2019deepred} proofed that DIP and its variations are very effective machines for handling various inverse problems, we have to figure out a stopping method before applying DIPs to real-world problems where human supervision is not available. Currently, DIPs stop when humans assess their outputs as good enough or reach their maximum iteration times \cite{ulyanov2018deep, mataev2019deepred}. The stopping method should output a measurement that indicates how well DIPs have reconstructed the interested image. So, the training algorithm can stop itself when the measurement reaches the maximum.

In this work, we propose a stopping method, namely Orthogonal Stopping Criterion (OSC), which adds a pseudo noise to the corrupted image and measure the pseudo-noise component in the recovered image of each iteration based on the orthogonality between signal and noise. The growth-rate derivate of the measurement will reach its maximum when DIPs start focusing on reconstructing the pseudo noise, which means the training should be stopped because DIPs resist "bad" solutions and descends much more quickly towards naturally-looking images \cite{ulyanov2018deep}. We use DIP as the baseline\footnote{https://github.com/DmitryUlyanov/deep-image-prior} and have demonstrated the performance of OSC for several problems such as denoising, super-resolution, inpainting.

%------------------------------------------------------------------------
\section{Methodology}
The inverse tasks such as denoising, super-resolution and inpainting can be expressed as energy minimization problem of equation~\eqref{orignal_loss}, where $E(x;x_0)$ is a task-dependent data term, $x_0$ is the noisy/low-resolution/occluded image, $x$ is the reconstructed image, and $R(x)$ is a regularizer \cite{ulyanov2018deep}. 

\begin{align}
x^* & = \arg\min_x E(x;x_0)+R(x)
\label{orignal_loss}
\end{align}

In this work, we handle the inverse tasks by equation~\eqref{pseudo_loss} where $pn$ is the pseudo noise and all $x$ that we get during the minimization are represented as set $\mathbb{X}$. The minimization is stopped according to equation~\eqref{OSC_loss}, where $O(x;pn)$ measures the growth-rate derivate of the pseudo-noise component in $x$. The pseudo-noise component is highly correlate to $pn$, because $pn$ is orthogonal to all components in $x_0$ including ground truth and corruptions. Since the reconstruction of $pn$ needs much more iterations than the naturally-looking image in $x_0$, the interested naturally-looking image will be reconstructed by DIP before the growth-rate derivate of the pseudo-noise component reaches its maximum, as long as the reconstruction difficulty of $pn$ is harder than the naturally-looking image and easier than (or equal to) other corruptions.

\begin{align}
& \min_x E(x;x_0+pn)+R(x)
\label{pseudo_loss}
\end{align}

\begin{align}
x^* & =\arg\max_x O(x;pn), x \in \mathbb{X}
\label{OSC_loss}
\end{align}

Given a series of reconstructed images $x_i, i=1,2,3, ...$, we get the pseudo noise component $e_i$ by equation~\eqref{noise_component}, where $N$ is the number of elements in $x_i$ and $i$ indicates the image which is reconstructed in the $i$th iteration.  

\begin{align}
e_i & = \frac{1}{N} \sum_{j=1}^{N} x_{i,j} \times pn_j
\label{noise_component}
\end{align}

\begin{figure}
\begin{center}
\includegraphics[width=1.0\linewidth]{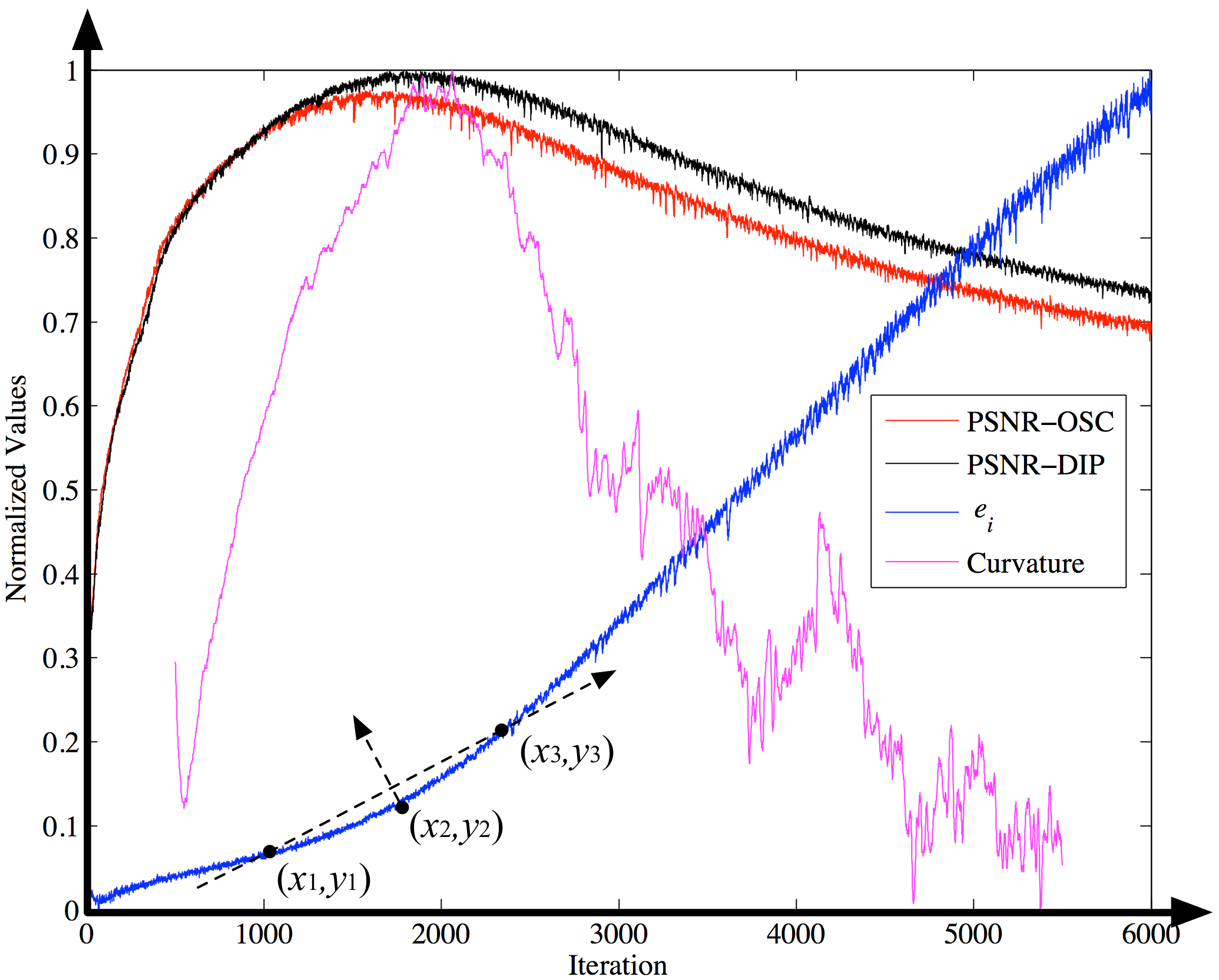}
\end{center}
   \caption{Example of the $e_i$ curve.}
\label{fig:error_curve}
\end{figure}

We get the index $i^*$ of the best image by equation~\eqref{curvature} where $C(e_i)$ finds the curvature of the $e_i$ curve. Figure~\ref{fig:error_curve} has shown the results of F16 denoising experiment, including the $e_i$ curve, the Peak Signal to Noise Ratio (PSNR) curve of DIP,  the PSNR curve of OSC, and curvature curves. All curves are normalized according to their own minimum and maximum except the PSNR curve of OSC which uses the minimum and the maximum of the DIP PSNR curve. The curvature curve records the growth-rate derivate of the $e_i$ curve. To get the curvature curve, for a specific $i$, we find 3 points on the $e_i$ curve to define the new coordinate system shown in dash line, which are $(x_1,y_1)$, $(x_2,y_2)$, $(x_3,y_3)$ where $x_1=i-H$,  $y_1=mean(e_{m_1})$, $m_1 \in [i-H-h,i-H+h]$, $x_2=i$, $y_2=mean(e_{m_2})$, $m_2 \in [i-h,i+h]$, $x_3=i+H$, $y_3=mean(e_{m_3})$, $m_3 \in [i+H-h,i+H+h]$. $H$ defines the length of $e_i$ curve for curvature calculation. $h$ is the averaging window size. After mapping the $e_i$ curve between $(x_1,y_1)$ and $(x_3,y_3)$ to the dash-line coordinate system, we fit a parabola to the curve and use the parameter of the quadratic item as the curvature at index $i$ which is an approximation of the growth-rate derivate of the pseudo-noise component. From Figure~\ref{fig:error_curve}, although the curvature-maximum PSNR is not the maximum one during the whole OSC iteration, it's close enough that the ratios of the curvature-maximum PSNR to the maximum one are more than 95\% in the most of our experiments. It is clear in Figure~\ref{fig:error_curve} that the existence of the pseudo noise will harm the maximum PSNR but it is insignificant. The OSC method has been listed in Algorithm~\ref{alg:OSC}.

\begin{align}
i^* & = \arg\max_i C(e_i)
\label{curvature}
\end{align}

\begin{algorithm}  
\scriptsize
  \renewcommand{\algorithmicrequire}{\textbf{Input:}}
  \renewcommand{\algorithmicensure}{\textbf{Output:}}
  \caption{Orthogonal Stopping Criterion}
  \label{alg:OSC}
  \begin{algorithmic}[1]
    \REQUIRE Corrupted image $x_0$, half window length $H$ for curvature calculation, $h$ as the half window length for averaging.
    \ENSURE The best index $i^*$ where the curvature-maximum PSNR has been reached. 
    \STATE  Generate a pseudo noise $pn$.
    \FORALL{$i$}    
    \STATE  Try to minimize equation~\eqref{pseudo_loss} and get a reconstructed image $x_i$
    \STATE  Get the measurement of pseudo noise component by equation~\eqref{noise_component}.
    \STATE  Calculate the curvature at $j=i-H-h$ by $C(e_j)$.
    \ENDFOR 
    \STATE \textbf{return} The index $i$ of the maximum curvature.
  \end{algorithmic}  
\end{algorithm}

%------------------------------------------------------------------------
\section{Experiments}
We tested OSC for denoising, super-resolution, and inpainting using same configration as \cite{ulyanov2018deep, ulyanov2018deepIJCV} \footnote{https://github.com/DmitryUlyanov/deep-image-prior}. In the following experiments, $H=200$, $h=20$, $pn$ is a 0 mean 1/25 standard deviation Gaussian pseudo noise for default. All OSC experiments are same as DIP's except the using of the pseudo noise. DIP experiments are stopped at suggested iteration or when PSNR reaches maximum. OSC experiments are stopped when the curvature reaches maximum.

\subsection{Denoising and generic reconstruction}
For denoising, we train OSC to minimize equation~\eqref{pseudo_loss} using $E(x;x_0+pn)=\|x-(x_0+pn)\|^2$ where $x$ is the reconstructed image, $x_0$ is a noisy observation. The pseudo-noise component is calculated by equation~\eqref{noise_component}.

\begin{figure*}[htb]
\begin{center}
\subfigure[Corrupted]{  
\includegraphics[width=4cm]{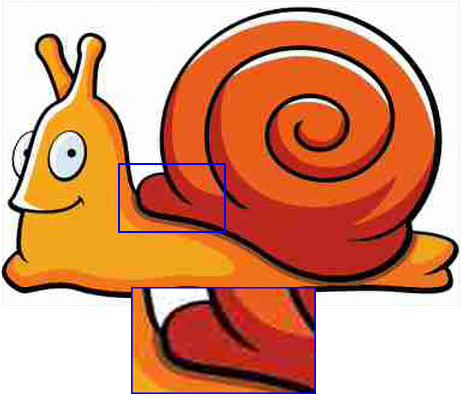}}
\hspace{0.1cm}    %每张图片中间空闲
\subfigure[DIP (2400 iterations)]{
\includegraphics[width=4cm]{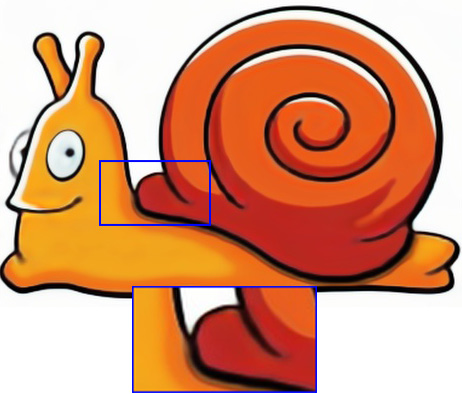}}
\hspace{0.1cm}
\subfigure[OSC]{
\includegraphics[width=4cm]{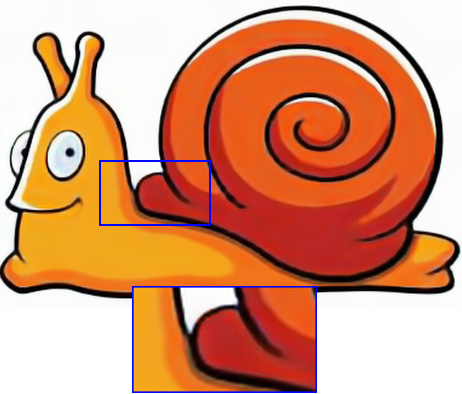}}
\end{center}
   \caption{Blind restoration of a JPEG-compressed image}
\label{fig:blind-JPEG}
\end{figure*}

Figure~\ref{fig:blind-JPEG} shows the restoration of a JPEG-compressed image where we repeat the experiment using DIP and OSC. Figure~\ref{fig:blind-JPEG} (b) is the image at the suggested stop iteration, (c) is obtained by OSC. The image automatically selected by OSC is better than the DIP result without the supervision of humans.

\begin{figure*}[htb]
\begin{center}
\subfigure[GT]{  
\includegraphics[width=3.2cm]{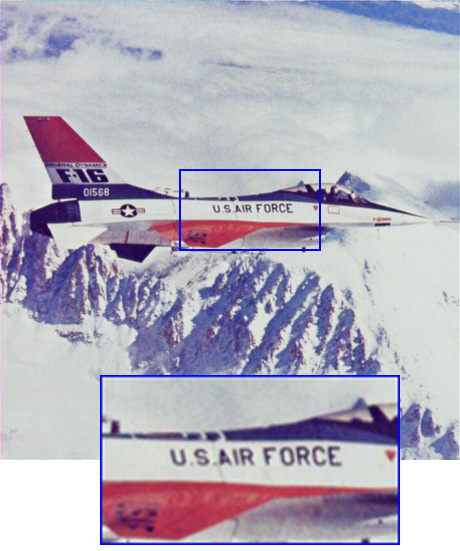}}
\hspace{0.05cm}    %每张图片中间空闲
\subfigure[Input]{
\includegraphics[width=3.2cm]{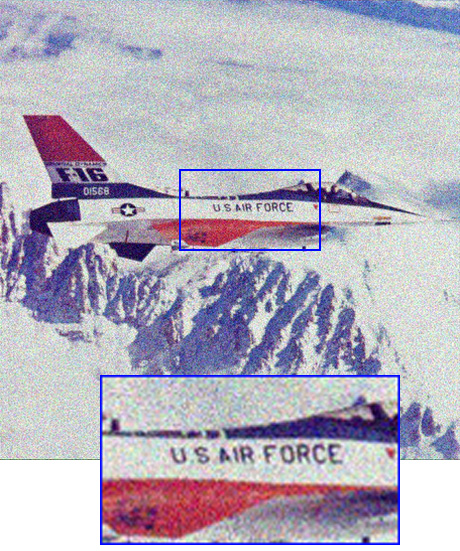}}
\hspace{0.05cm}
\subfigure[DIP (1804 iterations)]{
\includegraphics[width=3.2cm]{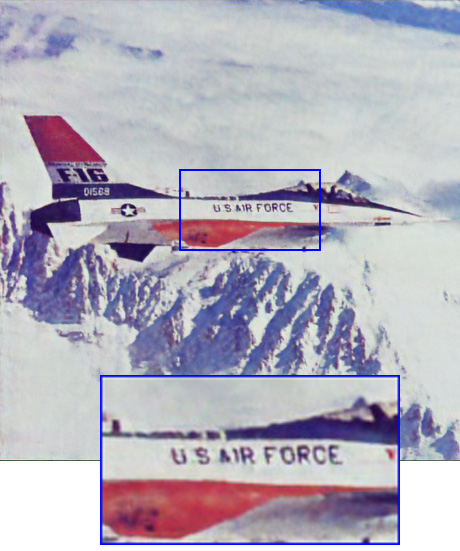}}
\hspace{0.05cm}
\subfigure[DIP (3000 iterations)]{
\includegraphics[width=3.2cm]{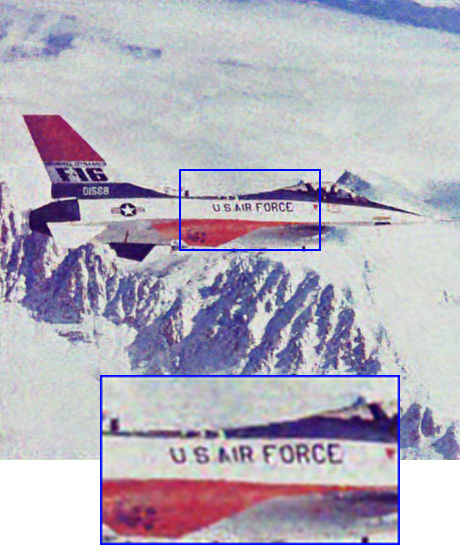}}
\hspace{0.05cm}
\subfigure[OSC]{
\includegraphics[width=3.2cm]{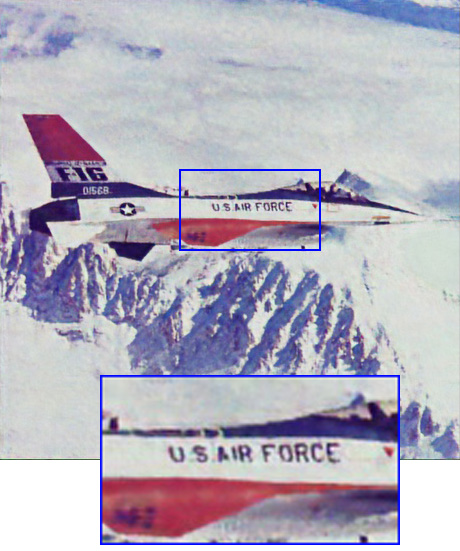}}
\end{center}
   \caption{Blind image denoising}
\label{fig:blind-denoising}
\end{figure*}

Figure~\ref{fig:blind-denoising} shows the denoising results of DIP and OSC, where (c) is selected by PSNR, (d) is the result of suggested iteration which is selected based on human inspection, (e) the result of OSC. The result of OSC is close to the PSNR-maximum image, and better than the suggested iteration.

We have done the denoising experiments on Kate, Snail, F16 images and the results are shown in table~\ref{tab:denoising}. The JPEG corrupted Snail image has been used as the ground truth. DIP (PSNR) gives out the maximum PSNR and DIP (Iteration) shows the PSNR of 3000 iteration which is the default value in the DIP code. The PSNR values of images selected by OSC are listed in the 4th row followed by the maximum PSNR that we have gotten during OSC iterating. Accuracy is the ratio of 4th row to 5th row. As shown in table~\ref{tab:denoising}, OSC results are comparable to DIP which needs the supervision of humans. The Max PSNR is close to DIP (PSNR) which means that the addition of pseudo noise has little influence on the noisy image reconstruction.

\begin{table}[htb]
\centering
\caption{Denoising results}
\begin{tabular}{cccc}
\hline
Image           & Kate    & Snail   & F16     \\ \hline
DIP (PSNR)      & 31.39   & 27.30   & 30.82   \\
DIP (Iteration) & 31.27   & 26.7    & 29.29   \\
OSC             & 30.73   & 26.44   & 29.80   \\
Max PSNR        & 31.19   & 27.42   & 30.33   \\
Accuracy       & 98.53\% & 96.43\% & 98.25\% \\ \hline
\end{tabular}
\label{tab:denoising} 
\end{table}

\subsection{Super-resolution}
For super-resolution, we train OSC to minimize equation~\eqref{pseudo_loss} using $E(x;x_0+pn)=\|d(x)-(x_0+d(pn))\|^2$ where $x$ is the reconstructed image, $x_0$ a down-sampled observation, $d(.)$ is Lanczos down sample method which is used by \cite{ulyanov2018deepIJCV}. We tested DIP and OSC on Set5 \cite{bevilacqua2012low} and Set14 \cite{zeyde2010single} with down scales 4 and/or 8. 

\begin{figure*}[htb]
\begin{center}
\subfigure[Original Image]{  
\begin{minipage}[b]{3.1cm}
\centering
\includegraphics[width=3.2cm]{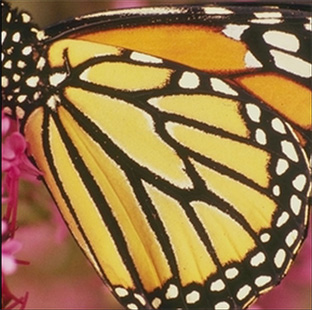} \\
\vspace{1pt}
\includegraphics[width=3.2cm]{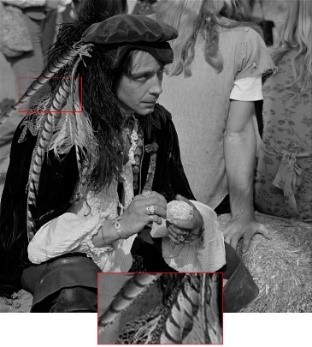}
\end{minipage}
}
\hspace{0.05cm}    %每张图片中间空闲
\subfigure[Bicubic]{  
\begin{minipage}[b]{3.1cm}
\centering
\includegraphics[width=3.2cm]{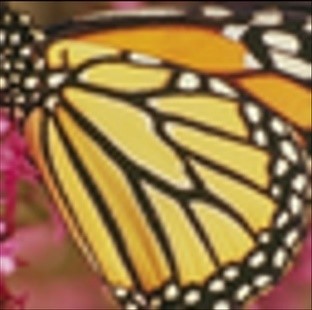} \\
\vspace{1pt}
\includegraphics[width=3.2cm]{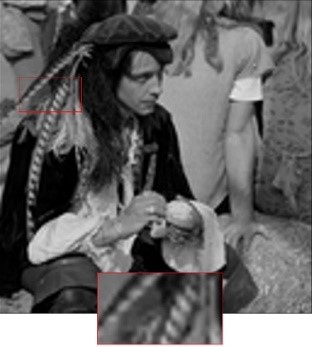}
\end{minipage}
}
\hspace{0.05cm}
\subfigure[DIP (Max PSNR)]{  
\begin{minipage}[b]{3.1cm}
\centering
\includegraphics[width=3.2cm]{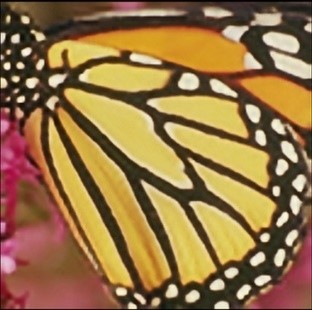} \\
\vspace{1pt}
\includegraphics[width=3.2cm]{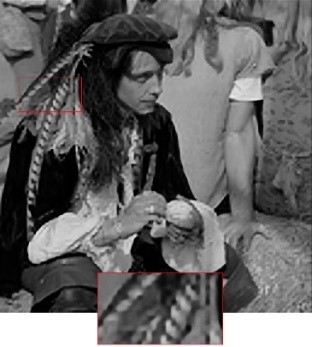}
\end{minipage}
}
\hspace{0.05cm}
\subfigure[DIP (2000 Iterations)]{  
\begin{minipage}[b]{3.1cm}
\centering
\includegraphics[width=3.2cm]{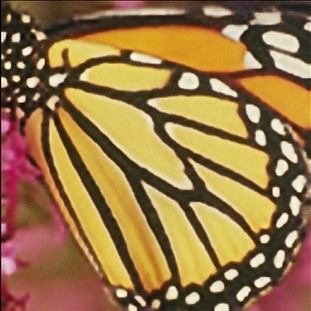} \\
\vspace{1pt}
\includegraphics[width=3.2cm]{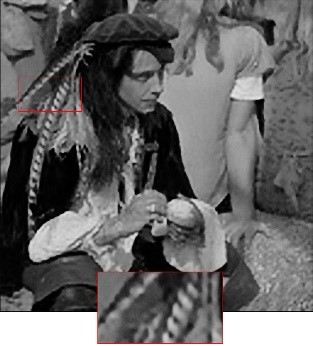}
\end{minipage}
}
\hspace{0.05cm}
\subfigure[OSC]{  
\begin{minipage}[b]{3.1cm}
\centering
\includegraphics[width=3.2cm]{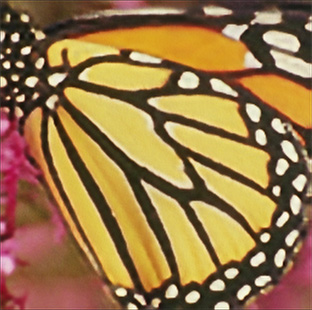} \\
\vspace{1pt}
\includegraphics[width=3.2cm]{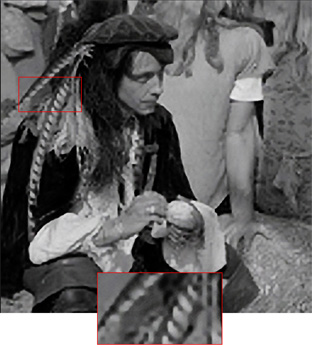}
\end{minipage}
}
\end{center}
   \caption{4x image super-resolution}
\label{fig:super-resolution}
\end{figure*}

Figure~\ref{fig:super-resolution} has shown the examples of 4x image super-resolution, where (c) is stopped at the PSNR-maximum point, (d) is stopped at the suggested 2000 iteration, (c) is generated by OSC. OSC results are very close to the PSNR-maximum version of DIP, which means that OSC has found the near-optimal solution for super-resolution inverse problem automatically.   

Table~\ref{tab:4xSet5} shows the 4x super-resolution results of Set5 where maximum PSNRs of DIP are in the 1st row, PSNRs of 2000 iterations DIP are in the 2nd row, OSC results are in the 3rd row followed by maximum PSNRs of OSC and accuracy in the last. Similarly, the results of 4x and 8x super-resolution on Set14 are shown in table~\ref{tab:4xSet14} and table~\ref{tab:8xSet14}. DIP was stopped at 8000 iterations in the 8x super-resolution experiment. From table~\ref{tab:4xSet5}, table~\ref{tab:4xSet14} and table~\ref{tab:8xSet14}, we believe that OSC is very good at finding optimal stopping iteration for super-resolution problems because the accuracy is higher than 95\% for all testing images.

\begin{table}[htb]
\centering
\caption{4x super-resolution on Set5}
\resizebox{0.5\textwidth}{!}{
\begin{tabular}{cccccc}
\hline
Image           & Baby    & Bird   & Butterfly  &Head &Woman    \\ \hline
DIP (PSNR)      & 30.66   & 30.33   & 24.93 & 28.90 & 27.50  \\
DIP (Iteration) & 29.78   & 29.63    & 24.69  & 28.42 & 26.93   \\
OSC             & 30.43   & 29.47   & 24.41     & 28.05   & 26.14    \\
Max PSNR       & 30.75   & 29.83   & 24.67     & 28.71   & 27.32  \\
Accuracy       & 98.96\% & 98.79\% & 98.95\%   & 97.70\% & 95.68\% \\ \hline
\end{tabular}}
\label{tab:4xSet5} 
\end{table}

\begin{table*}[htb]
\centering
\caption{4x super-resolution on Set14}
\resizebox{\textwidth}{!}{
\begin{tabular}{ccccccccccccccc}
\hline
Image     & Baboon  & Barbara & Bridge  & Coastguard & Comic   & Face    & Flowers & Foreman & Lenna   & Man     & Monarch & Pepper  & Ppt3    & Zebra    \\ \hline
DIP (PSNR)     & 20.45   & 23.95   & 23.25   & 24.56      & 21.00   & 29.00   & 25.04   & 28.29   & 29.56   & 25.19   & 29.48   & 28.50   & 22.99   & 24.49   \\
DIP (Iteration) & 20.27   & 23.78   & 23.13   & 24.39      & 20.86   & 28.38   & 24.48   & 27.83   & 29.03   & 24.81   & 28.74   & 27.87   & 22.67   & 24.07   \\
OSC             & 20.36   & 22.75   & 23.22   & 24.31               & 20.37   & 28.74   & 23.11   & 27.46            & 28.32     & 24.32     & 29.12     & 28.31     & 22.88     & 24.09   \\
Max PSNR       & 20.37   & 23.89   & 23.25   & 24.34           & 20.98   & 28.82   & 24.85   & 27.65            & 29.23     & 25.12     & 29.28     & 28.40     & 23.23     & 24.28   \\
Accuracy      & 99.95\% & 95.23\% & 99.87\% & 99.88\%    & 97.09\% & 99.72\% & 93.00\% & 99.31\% & 96.89\% & 96.82\% & 99.45\% & 99.68\% & 98.49\% & 99.22\% \\ \hline
\end{tabular}}
\label{tab:4xSet14} 
\end{table*}

\begin{table*}[htb]
\centering
\caption{8x super-resolution on Set14}
\resizebox{\textwidth}{!}{
\begin{tabular}{ccccccccccccccc}
\hline
Image     & Baboon  & Barbara & Bridge  & Coastguard & Comic   & Face    & Flowers & Foreman & Lenna   & Man     & Monarch & Pepper  & Ppt3    & Zebra    \\ \hline
DIP (PSNR)     & 19.38   & 22.35   & 21.13   & 22.60      & 18.42   & 27.29   & 21.36   & 24.08   & 26.68   & 22.55   & 23.96   & 25.96   & 18.78   & 19.62 \\
DIP (Iteration) & 19.36   & 22.33   & 21.10   & 22.58      & 18.37   & 27.10   & 21.34   & 23.89   & 26.59   & 22.52   & 23.92   & 25.86   & 18.67   & 19.58   \\
OSC             & 19.05     & 21.46     & 20.95      & 22.38       & 18.03      & 25.63     & 20.99     & 23.08     & 25.67     & 21.41      & 23.72     & 24.98     & 18.69      & 18.85  \\
Max PSNR   & 19.36     & 22.34     & 21.07      & 22.66       & 18.39      & 27.18     & 21.39     & 23.80     & 26.57     & 22.47       & 23.92    & 25.93     & 18.76      & 19.62    \\
Accuracy      & 98.40\% & 96.06\% & 99.43\% & 98.76\%    & 98.04\% & 94.30\% & 98.13\% & 96.97\% & 96.61\% & 95.28\% & 99.16\% & 96.34\% & 99.63\% & 96.08\% \\ \hline
\end{tabular}}
\label{tab:8xSet14} 
\end{table*}

\subsection{Inpainting}
For inpainting, we train OSC to minimize equation~\eqref{pseudo_loss} using $E(x;x_0+pn)=\|(x-(x_0+pn)) \odot m\|^2$ where $x$ is the reconstructed image, $x_0$ is a corrupted observation, $\odot$ is Hadamard's product, $m \in \{0,1\}^{F \times W} $ is a binary mask of the missing pixels in $x_0$, $F$ is the height and $W$ is the width of the image. $pn$ is the pseudo noise generated by equation~\eqref{inpaintingPN}, where $a$, $b$, $c$ are channel, row, column index respectively,  $A_b \in \mathbb{U}(\frac{1}{50}, \frac{1}{25})$, $\theta_b \in \mathbb{U}(\frac{\pi \times 20}{W}, \frac{\pi \times 40}{W})$, $\mathbb{U}$ is the random uniform distribution. The pseudo-noise component is calculated by equation~\eqref{inpainting_noise_component}, where $L=\sum_{j=1}^{N} m_j$, $N$ is the number of elements in the image.

\begin{align}
pn_{a,b,c} & =A_b \times \sin(\theta_b \times c)
\label{inpaintingPN}
\end{align}

\begin{align}
e_i & = \frac{1}{L} \sum_{j=1}^{N} x_{i,j} \times pn_j \times m_j
\label{inpainting_noise_component}
\end{align}

\begin{figure*}[htb]
\begin{center}
\subfigure[Original Image]{  
\includegraphics[width=0.19\linewidth]{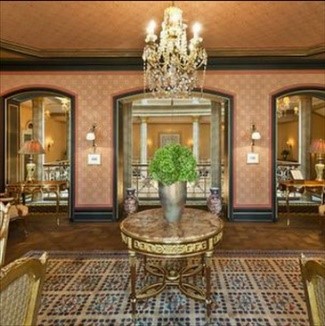}}
%\hspace{0.05cm}    %每张图片中间空闲
\subfigure[Corrupted Image]{
\includegraphics[width=0.19\linewidth]{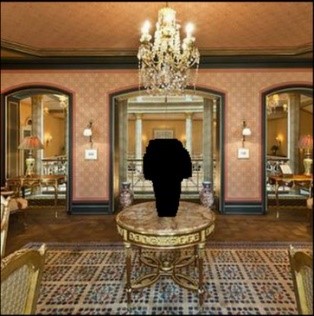}}
%\hspace{0.05cm}
\subfigure[DIP (Max PSNR)]{
\includegraphics[width=0.19\linewidth]{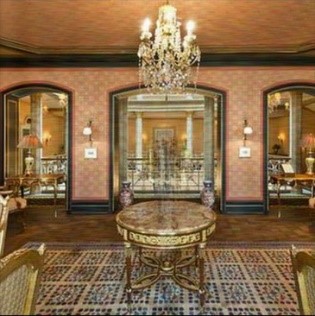}}
%\hspace{0.05cm}
\subfigure[DIP (5000 iterations)]{
\includegraphics[width=0.19\linewidth]{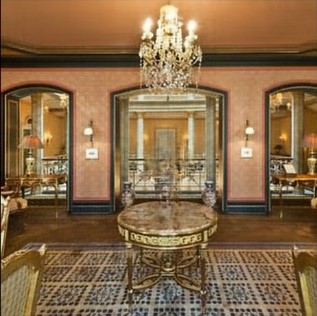}}
%\hspace{0.05cm}
\subfigure[OSC]{
\includegraphics[width=0.19\linewidth]{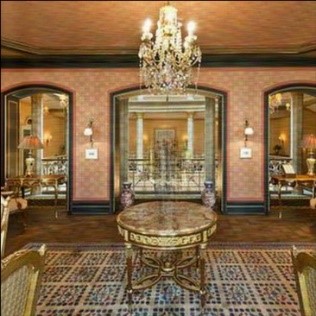}}
\end{center}
   \caption{The recovery of image Vase}
\label{fig:region-inpainting1}
\end{figure*}

\begin{figure*}[htb]
\begin{center}
\subfigure[Original Image]{  
\includegraphics[width=0.49\linewidth]{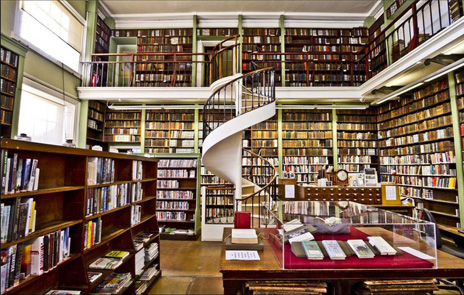}}
%\hspace{0.05cm}    %每张图片中间空闲
\subfigure[Corrupted Image]{
\includegraphics[width=0.49\linewidth]{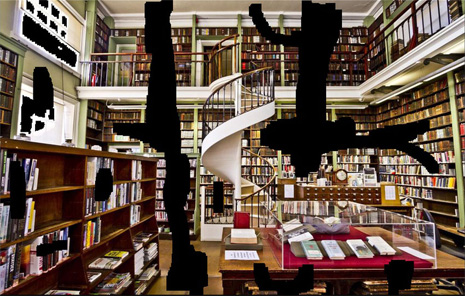}}
%\hspace{0.05cm}
\subfigure[DIP (Max PSNR)]{
\includegraphics[width=0.49\linewidth]{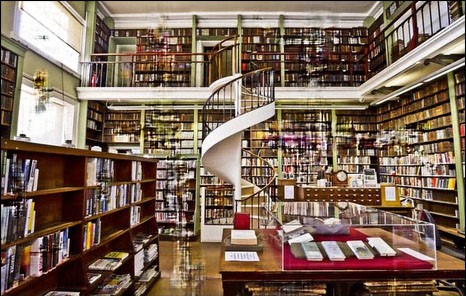}}
%\hspace{0.05cm}
\subfigure[DIP (5000 iterations)]{
\includegraphics[width=0.49\linewidth]{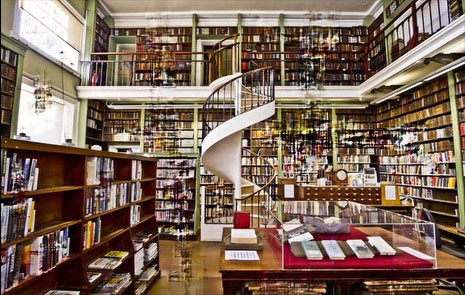}}
%\hspace{0.05cm}
\subfigure[OSC]{
\includegraphics[width=0.49\linewidth]{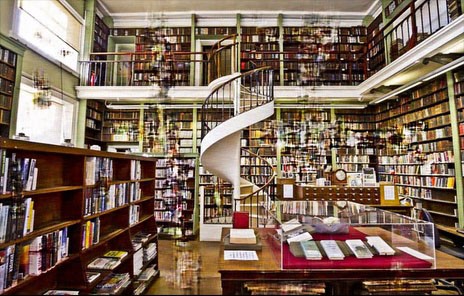}}
\end{center}
   \caption{The recovery of image Library}
\label{fig:region-inpainting2}
\end{figure*}

\begin{figure*}[htb]
\begin{center}
\subfigure[Original Image]{  
\includegraphics[width=0.24\textwidth]{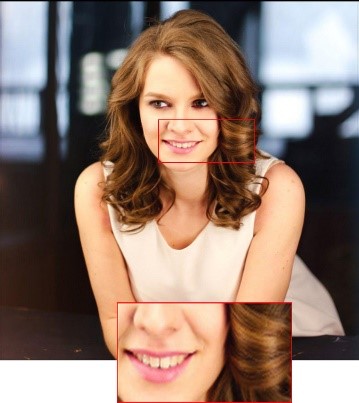}}
%\hspace{0.05cm}    %每张图片中间空闲
\subfigure[Corrupted Image]{
\includegraphics[width=0.24\linewidth]{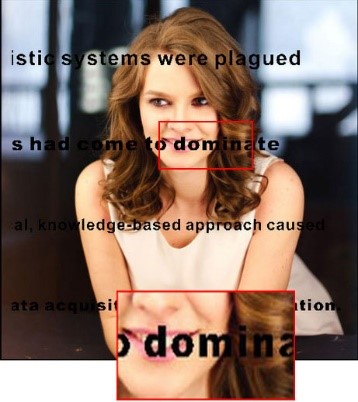}}
%\hspace{0.05cm}
\subfigure[DIP (Max PSNR)]{
\includegraphics[width=0.24\linewidth]{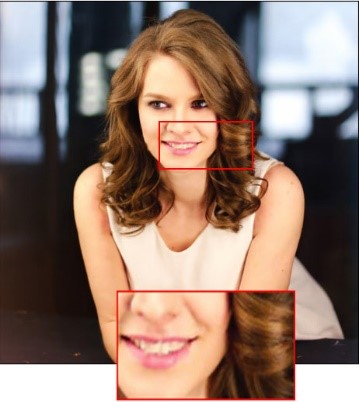}}
%\hspace{0.05cm}
\subfigure[OSC]{
\includegraphics[width=0.24\linewidth]{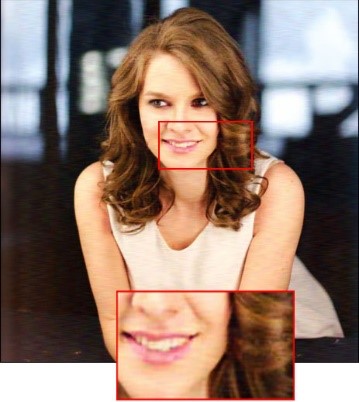}}
%\hspace{0.05cm}
\subfigure[Original Image]{  
\includegraphics[width=0.24\linewidth]{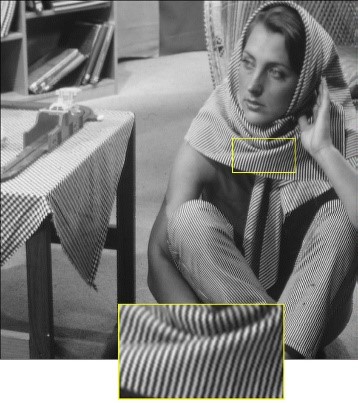}}
%\hspace{0.05cm}    %每张图片中间空闲
\subfigure[Corrupted Image]{
\includegraphics[width=0.24\linewidth]{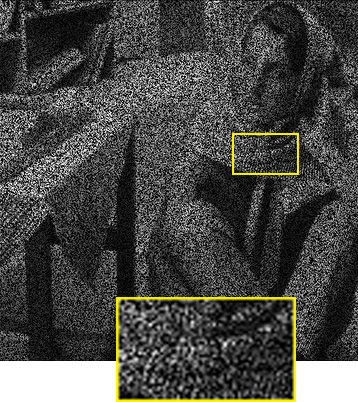}}
%\hspace{0.05cm}
\subfigure[DIP (Max PSNR)]{
\includegraphics[width=0.24\linewidth]{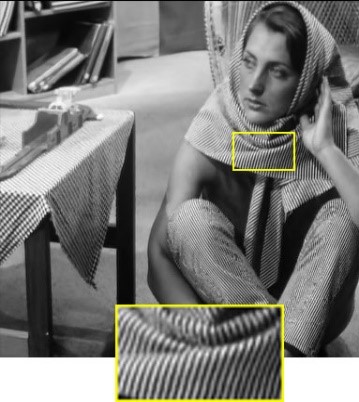}}
%\hspace{0.05cm}
\subfigure[OSC]{
\includegraphics[width=0.24\linewidth]{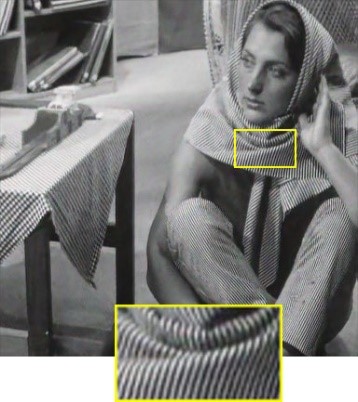}}
\end{center}
   \caption{Two inpainting approaches}
\label{fig:inpainting}
\end{figure*}

Figures~\ref{fig:region-inpainting1} and~\ref{fig:region-inpainting2} shows the results of regional recovery. Figure~\ref{fig:inpainting} shows the results of two inpainting approaches. Table~\ref{tab:inpainting} lists PSNRs of the experiments. The OSC results are very close to DIP maximum PSNR which means the OSC method is fully capable of finding optimal stopping iteration automatically.

\begin{table}[htb]
\centering
\caption{PSNRs of inpainting experiments}
\resizebox{1\linewidth}{!}{
\begin{tabular}{cccccc}
\hline
Image           & Kate    & Library & Vase    & Barbara\\ \hline
DIP (PSNR)      & 40.19   & 19.22   & 29.14 & 31.91 \\
DIP (Iteration) & 39.14   & 19.08   & 27.76    & 30.90\\
OSC              & 33.74    & 18.64      & 28.67     & 30.97\\
Max PSNR    & 35.33    & 18.72      & 28.71     & 31.07 \\
Accuracy       & 95.50\% & 99.57\% & 99.86\% & 99.68\% \\ \hline
\end{tabular}}
\label{tab:inpainting} 
\end{table}

%------------------------------------------------------------------------
\section{Conclusion}
In this work, we have developed Orthogonal Stopping Criterion (OSC) which can endow Deep Image Prior (DIP) the power of automation. The automatic stopping mechanic is essential to DIP in real-world applications because the Peak Signal to Noise Ratio (PSNR) and human supervision are both unavailable or hard to reach. By adding pseudo noise to the corrupted image, OSC can find the near-optimal result automatically which is very close to the one with maximum PSNR in our experiments. Additionally, the pseudo noise has little influence on the maximum PSNR which has been verified by the experiments. The ratios of OSC PSNR to the maximum are higher than 95\% in 38 out of 40 experiments. Many of them are even higher than 99\%. Although, the results of DIP are comparable to OSC, they are selected based on PSNR or human inspection. In all, we believe that OSC is an indispensable part of DIP-based single image inverse systems.

{\small
\bibliographystyle{ieee_fullname}
\bibliography{egbib}
}

\end{document}